\documentclass[letterpaper, 10 pt, conference]{ieeeconf}  

\IEEEoverridecommandlockouts                              

\usepackage{amssymb}
\usepackage{mathtools}
\usepackage{booktabs}
\usepackage{subcaption}
\usepackage{caption}
\usepackage{wrapfig}
\usepackage[capitalize,noabbrev]{cleveref}

\usepackage{footnote}
\usepackage{xcolor}
\usepackage{url}

\usepackage{selectp}

\definecolor{ms_red}{RGB}{167, 22, 34}

\definecolor{rebuttaltext}{RGB}{84, 11, 14}
\definecolor{rebuttaltext2}{RGB}{17, 79, 137}
\def\shownotes{1}  
\ifnum\shownotes=1
\newcommand{\authnote}[2]{{$\ll$\textsf{\footnotesize #1 notes: #2}$\gg$}}
\else
\newcommand{\authnote}[2]{}
\fi

\newcommand{\mpinetname}[0]{M$\pi$Nets }
\newcommand{\aitstar}[0]{AIT$^{*}$}

\title{\LARGE \bf
Cascaded Diffusion Models for Neural Motion Planning
}

\author{Mohit Sharma$^{1}$, Adam Fishman$^{2}$, Vikash Kumar$^{1}$, Chris Paxton$^{3}$ and Oliver Kroemer$^{1}$
\thanks{$^{1}$Robotics Institute,  Carnegie Mellon University, Pittsburgh, PA, USA}
\thanks{$^{2}$University of Washington, Seattle, WA, USA}
\thanks{$^{3}$Hello Robot Inc, USA}
\thanks{Correspondence to \texttt{mohitsharmacmu@gmail.com}}
}

\begin{document}

\maketitle
\thispagestyle{empty}
\pagestyle{empty}

\maketitle

\begin{abstract}
Robots in the real world need to perceive and move to goals in complex environments without collisions. Avoiding collisions is especially difficult when relying on sensor perception and when goals are among clutter. Diffusion policies and other generative models have shown strong performance in solving \textit{local} planning problems, but often struggle at avoiding all of the subtle constraint violations that characterize truly challenging global motion planning problems.
In this work,
we propose an approach for learning global motion planning using diffusion policies, allowing the robot to generate full trajectories through complex scenes and reasoning about multiple obstacles along the path. Our approach uses cascaded hierarchical models which unify global prediction and local refinement together with online plan repair to ensure the trajectories are collision free. 
Our method outperforms $(\approx 5\%)$ a wide variety of baselines on challenging tasks in multiple domains including navigation and manipulation.

\end{abstract}


\section{Introduction}

%
A key requirement for useful robots is that they can generalize motions to new environments. While classical motion planning algorithms often show good generalization~\cite{kuffner2000rrt}, they require privileged information (e.g., full scene geometry) about their world; this has led to interest in neural motion planning approaches which can operate off of raw sensor data~\cite{qureshi2019motion,fishman2023motion,danielczuk2021object,qureshi2021nerp,goyal2022ifor}, and leverage large-scale behavior cloning to guide sampling~\cite{pomerleau1988alvinn, qureshi2019motion, fishman2023motion}. However, neural motion planning approaches often struggle at generalizing to the challenging, cluttered environments in which traditional motion planners excel.
This limitation is because learned approaches fail to satisfy \textit{all} of the many constraints necessary for a trajectory to be successful for a high-dimensional multi-modal planning problem.


Fortunately, there exists a class of models that are well suited to this sort of multimodal, high-dimensional data: diffusion models~\cite{karras2022elucidating, chi2023diffusion,wang2022diffusion,liu2022structdiffusion}.
Diffusion models have become extremely popular for image generation tasks due to their expressivity and controllability \cite{rombach2022high, ramesh2021zero, saharia2022photorealistic, ho2022classifier, zhang2023adding}.
Recently, they have been used for motion generation within robotics:  predicting both
low-level visuo-motor control outputs \cite{reuss2023goal, chi2023diffusion}
and  plan outputs \cite{janner2022planning, wang2022diffusion, carvalho2023motion, liu2022structdiffusion}.
Both sets of motion generation tasks involve multi-modality (either from human demonstrations or motion planning) and require some controllability (i.e., outputting low cost, smooth, and safe motions).

In this work, we focus on the ability of diffusion models to generate 
near-optimal goal-reaching motion plans in diverse, complex environments directly from raw sensory data.
Generating near-optimal long-horizon plans that successfully reach goal states requires both \emph{global} and \emph{local} information. 
Global information
determines the overall plan structure, as distant obstacles can influence which actions or paths need to be taken earlier.
On the other hand, local information around the current configuration and nearby obstacles is important to avoid violating constraints like obstacle collisions and self-collisions.

Previous learning based approaches for plan generation \cite{qureshi2019motion, fishman2023motion} use
less expressive model classes like multi-layer perceptrons, which struggle to model multi-modal 
distributions~\cite{bishop1994mixture}.
While more recent approaches \cite{johnson2021motion, johnson2023learning} use more expressive architectures like Transformers~\cite{vaswani2017attention}, they still rely on a classical planner to generate the final plans.
Some recent approaches have successfully applied diffusion
models for plan generation~\cite{janner2022planning, wang2022diffusion, carvalho2023motion, saha2023edmp}.
These approaches forgo using classical planners for plan generation.
However, these works either only use fixed 3D environments 
\cite{janner2022planning, wang2022diffusion},
or rely on privileged information (i.e. full scene geometry and collision gradients) to perform inference-time adaptation \cite{carvalho2023motion, saha2023edmp}.
By contrast, we propose a novel framework to learn in large-scale diverse environments and plan directly from raw perceptual input without access to any privileged information.

In this work, we propose a hierarchy of cascaded diffusion models for plan generation.
Each diffusion model in our hierarchy operates at a 
different temporal resolution. 
This approach allows us to output low cost, long horizon plans while using a fixed planning horizon for each model.
Further, the cascaded approach allows 
the lower-level models to use the high-level diffusion model's output as guidance to produce locally-consistent plans.
Finally, we propose a plan refinement approach which allows us to handle very challenging environments.
Overall, our contributions include:
\begin{itemize}
   \item A novel cascaded diffusion framework for plan generation in complex and challenging manipulation settings using noisy raw perceptual input only. 
   \item A plan refinement strategy to improve the generalization and robustness of our approach. 
   \item Comprehensive evaluations across 3 different experimental domains, including 2D navigation and 7D manipulator planning domains with  both local and global planning.
\end{itemize}

\begin{figure*}[t]
    \centering
    \includegraphics[width=0.8\textwidth]{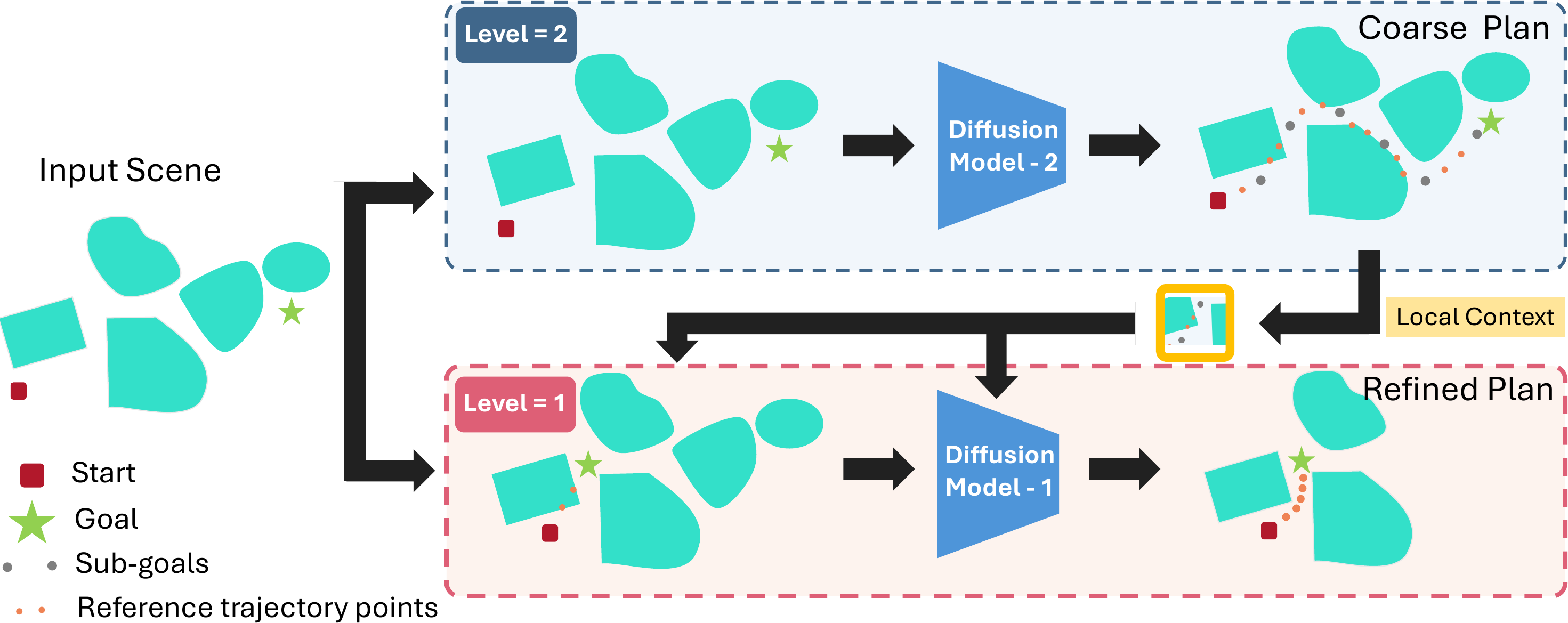}
    \caption{\footnotesize{Overview of our cascaded diffusion model approach. The higher level model generates coarse plan as sub-goals and reference points. The lower level model uses them as input to output a plan that satisfies local constraints.}}
    \label{fig:overview}
\end{figure*}

\section{Related Work}
\label{sec:related-work}


\textbf{Robot Motion Planning} is a large field of robotics, which can very broadly be categorized into discrete search based planners (such as A*, MHA*)\cite{aine2016multi, hansen2007anytime}, continuous sampling based planners (such as RRT*, PRMs) \cite{kuffner2000rrt, lavalle2001randomized, kavraki1996probabilistic}, and trajectory optimization (such as CHOMP, TrajOpt)~\cite{chomp, schulman2014motion, sundaralingam2023curobo, kalakrishnan2011stomp}. 
Most robotic manipulation planners aim to achieve global optimality by using a known low-dimensional model of the environment \cite{strub2020adaptively, gammell2015batch, lavalle2006planning}.
While these planning techniques have found widespread use in robotics, they can often struggle in some scenarios. For instance, while RRT*-based planners guarantee asymptotic optimality, they can sometimes produce erratic and sub-optimal plans given a restricted planning time. This problem becomes especially pronounced for high-dimensional robot systems.

\textbf{Neural Motion Planning}
approaches aim to address the above shortcomings of classical motion planning techniques.
These works can be divided into two broad categories. First, works which use learning based techniques within the planning algorithm. Examples of such works include using learned samplers (e.g. learning sampling for RRT) \cite{ichter2018learning, chamzas2022learning, chamzas2021-learn-sampling, johnson2023learning} or learned heuristics \cite{bhardwaj2017learning}. 
Second, other works focus on directly using learning techniques to generate robot plans.
These include Neural-A* \cite{yonetani2021path} and Value-Iteration-Networks \cite{tamar2016value}, which try to mimic the planning algorithm structure within the neural network.
Alternatively, recent works also use imitation learning \cite{janner2022planning, qureshi2019motion, fishman2023motion} or reinforcement learning \cite{jurgenson2019harnessing, jurgenson2020sub} to learn robot plans directly from expert trajectories or interactions respectively.
Our work lies in this latter category, i.e., we use \emph{large-scale} imitation learning to train a model to generate motion plans. However, different from prior works, we propose a hierarchical imitation learning objective and show that it exceeds other prior choices used in the literature.

\textbf{Diffusion Models for Robotics}
have recently emerged as a popular choice to model complex multi-modal distributions and have seen rapid adoption within the robotics community \cite{janner2022planning, reuss2023goal, chi2023diffusion, liu2022structdiffusion, urain2023se, mishra2023generative}.
Diffusion models have shown to better model multi-modality and thus have been used to generate motion plans \cite{janner2022planning, carvalho2023motion} as well as policies \cite{reuss2023goal, chi2023diffusion} for contact-rich manipulation tasks.
Their use for generating motion plans is due to their improved ability to model complex multi-modal distributions as well as their use in gradient based planning \cite{saha2023edmp}.

Our work also focuses on using diffusion models for generating motion plans. 
Among prior works that use diffusion models for generating plans, initial works focused on generating plans for a fixed environment configuration \cite{janner2022planning, wang2022diffusion, carvalho2023motion}.
By contrast, more recent works have focused on generating plans for a large variety of scenes in complex settings \cite{saha2023edmp}.
Similar to these previous works we focus on a large set of challenging environment configurations. However, prior works often use a diffusion models in a naive non-hierarchical  manner for generating motion plans, which we show is insufficient for generalization to complex scenes. 
By contrast, we propose a hierarchical diffusion \cite{chen2020simple, huang2024subgoal} approach which focuses on both long-term planning as well as generating short-term collision free trajectories.
Most similar to our approach is \cite{chen2024simple} which uses independently 
trained hierarchical diffusion models on fixed environment configurations.
However, as we show empirically, independently trained hierarchical models can perform sub-optimally on more complex problems.


\section{Preliminaries}
\label{sec:preliminaries}

We consider robot workspace $\mathit{W}$, which is sampled from a large set of environment configurations. Each workspace $\mathit{W}$ consists of a set of obstacles which generate the C-space $X_\text{obs} \in \mathit{C}$. 
Each workspace also has a start $q_0$ and goal $q_g$ configuration and the planning problem is to quickly find a collision free path $\mathit{X}_\text{path}$ from $q_0$ to $q_g$, which does not lie in $\mathit{X}_\text{obs}$.
To achieve this goal, the robot only observes some discrete points randomly sampled on each obstacle. 
For each training workspace $\mathit{W}$ we also generate an optimal low-cost plan $\mathit{X}_\text{path}$ using a classical motion planner \cite{sucan2012the-open-motion-planning-library}.

\textbf{Diffusion Models} are generative models that use an iterative denoising process to 
map between two different distributions.
Most commonly we use Gaussian noise and repeatedly add it to our data distribution to get the source distribution. 
Formally, denoting $x$ as our event space, we get 
$q\left( x_t \vert x_{t -1 } \right) = \mathcal{N}\left( x_t; \sqrt{\alpha_t}x_{t-1}, (1 - \alpha_t)\mathbf{I} \right)$, 
where $x_0 \sim p_\text{data}$ is a sample from the true data distribution and $x_t$ is a sample after $t$ steps of adding Gaussian noise
(this is different from the trajectory timestep).
We use a fixed schedule over $\alpha_t$ to add increasingly large amount of noise to convert true data samples $x_0$ to isotropic Gaussian noise $x_T$.
Diffusion models are tasked with learning the reverse process, i.e. mapping from isotropic Gaussian noise to the true data distribution.
Formally, they are trained to approximate 
$p_\theta\left( x_{t-1} \vert x_t \right) = \mathcal{N}\left( x_{t-1}; \mu_\theta (x_t, t), \Sigma_t \right)$. 
Recent works \cite{janner2022planning, wang2022diffusion} have
shown how the diffusion process can also be used for plan generation, specifically to generate a diverse set of plans in fixed environments.
They use a \emph{fixed planning horizon} for the given task and use
temporal U-Nets \cite{unet} with one-dimensional temporal convolutions to 
generate temporally local plans with each convolution having a fixed kernel size.

\section{Approach}

Using learning based approaches for plan generation in complex environments requires the model to generate near-optimal plans while not violating any constraints (e.g. collision constraints, joint limits, etc.).
This is challenging because optimal plans require the model to consider \emph{global information} about the entire environment and goals. However, ensuring adherence to constraints necessitates a focus on low-level \emph{local information} in the immediate surroundings.
To address this challenge, we propose a hierarchical approach using cascaded diffusion models~\cite{ho2022cascaded}. 


Assuming a cascading hierarchy with $K$ levels, we first sequentially generate appropriate trajectories (or sub-goals) for training each level. 
The entire input trajectory $\tau = [q_0, q_1, \cdots, q_T] $ is used for training level $1$. 
At level 2, we sub-sample states from $\tau$ with some fixed frequency $n_2$ to get a new sub-sampled trajectory
$\tau^2 = \left[ q_0, q_{n_2}, q_{2 \times n_2}, \cdots \right]$.
Importantly, we use multiple start configurations from $\tau$ to get multiple distinct trajectories at level 2 from the same base trajectory. We generate training trajectories at higher levels in the same manner.


\begin{figure*}[bt]
    \centering
    \includegraphics[width=0.8\textwidth]{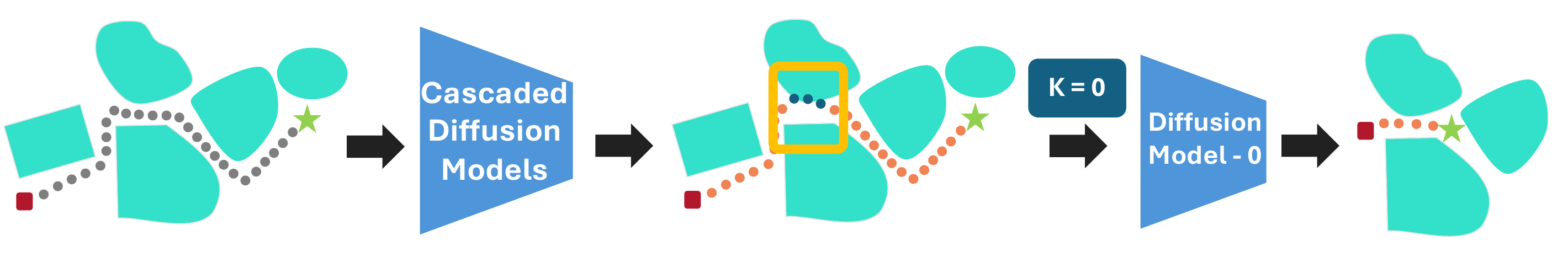}
    \caption{\footnotesize{We use the lowest level model in our cascaded hierarchy of diffusion models to refine paths. This diffusion model operates locally with the same planning horizon as the model directly above it.}}
    \label{fig:path-patching}
    \vspace{-8pt}
\end{figure*}

\vspace{-4pt}
\subsection{Hierarchical Cascaded Diffusion Model}
We propose to use a cascaded diffusion architecture \cite{ho2022cascaded} to model the hierarchy of trajectories generated above. 
Specifically, we begin with a diffusion model at the highest level $K$,
which generates a coarse plan to reach the goal from a given start state.
Importantly, since the trajectory at level $K$ is sub-sampled and is thus short in length we can use a fixed short planning horizon to model the entire trajectory.
Using a short planning horizon at higher level $K$ allows the learned planner to focus on producing globally optimal trajectories.

Once we have generated trajectories at level $K$, we refine them at lower levels starting at $K - 1$.
In the following discussion,
for ease of exposition, we focus on the case where $K = 2$.
The diffusion model at level $K - 1 = 1$ also uses a 
small fixed planning horizon.
However, instead of focusing on the entire trajectory (i.e. global optimality),
this diffusion model focuses on local constraints while conditioning
on the states generated by the higher level diffusion model.
This conditioning on the output of the higher-level diffusion happens in two ways.
First, states generated by the higher level diffusion model, which are further away from the current state,
are directly used as sub-goals for the low-level diffusion model.
However, solely relying on the low-level diffusion model to generate intermediate states is inefficient and,
as we show empirically, sub-optimal.
To avoid this we ensure that the high-level diffusion model ($K=2$) also generates sparse intermediate states between sub-goals.
We densify these sparse intermediate states using 
linear interpolation and use them as latent inputs to the lower-level diffusion model.
These intermediate states provide better conditioning to the lower-level diffusion model while ensuring faster learning.
We refer to these intermediate states as \emph{reference trajectory points} for our low-level diffusion model.

Fig~\ref{fig:overview} shows an overview of our cascaded approach.
As shown in Fig~\ref{fig:overview}, we begin with our coarse (high-level) model (at level $K$) and generate coarse plans from it, i.e, $p\left(x_0=\tau^K \vert o, q_0, q_g \right)$,
where $o$ denotes our observation (an image or point cloud) and  $q_0, q_g$ denote our start and goal configurations respectively.
We provide architecture details in Section~\ref{subsec:train-details}.
Following this, the next lower-level model (i.e. at level $K-1$), uses the output 
from the higher level model (both the sub-goal and the reference trajectory points) 
to further refine the trajectory locally.
Jointly, after $K$ levels of hierarchy we get the final plan,
    $p\left(x_0 = \tau \vert o, q_0, q_g\right) = p\left(x = \tau^{K} \vert o, q_0, q_g\right) \prod_{k=1}^{K-1} p\left(x = \tau^{K - k} \vert o, q_0, q_g, \tau^{K-k + 1}\right)$.

\subsection{Patching Plans}
The above coarse-to-fine formulation outputs close to optimal plans to reach goal configurations without violating local constraints.
However, for complex environments, the plans generated from high-level models can sometimes be inaccurate with higher-level models producing intermediate and sub-goal states which violate constraints. 
While our lower-level model can locally repair plans where intermediate states are inaccurate, it fails to recover plans where the sub-goals produced by the higher level model are invalid.

To alleviate this issue, we use a final plan refinement strategy wherein we directly use the output
of the lowest cascaded diffusion model.
For this plan refinement, we first find states  that violate constraints such as obstacle collisions or self-collisions.
Most states that violate constraints are often clustered together around complex obstacle configurations.
Thus, we find connected segments of these violating states.
With the connected segments we get a \emph{sequence} of invalid states and thus \emph{invalid sub-trajectories}.
To patch these invalid sub-trajectories, we first find valid states closest to the edges of the sub-trajectories.
Given these valid states, 
we can now in principle re-run our cascaded diffusion approach beginning from the high level model.
However, for all environments, we find that constraint violation is localized to small regions.
Hence, we only re-run the lowest level diffusion model which uses a small planning horizon for local correction.
We use the invalid set of states as the reference trajectory for this diffusion model.
Fig~\ref{fig:path-patching} presents an overview of our approach.
While plan refinement can be run recursively any number of times, for faster planning and for better comparisons, we only run it once.

\vspace{-8pt}
\subsection{Training and Architecture Details}
\label{subsec:train-details}
We train our hierarchy of cascaded diffusion models sequentially. 
We initially pre-train our highest level ($K$) diffusion model which does not rely on any sub-goals or reference trajectories. 
This model generates the first sparse set of sub-goals and intermediate waypoints for reference.
We then use this pretrained model to bootstrap learning for the next level of the hierarchy.
Thus, we are able to \emph{efficiently} train all models of our joint cascaded model.
In our formulation, low-level diffusion models use reference trajectories from higher level ones.
To allow for stable training, we ensure that our low-level diffusion model is robust to noise in the reference trajectory.
We achieve this by creating an appropriate set of data augmentations to train our low-level diffusion models.
The main augmentations we consider during training include using noisy diffusion outputs from the high-level diffusion policy
and conditioning on noise-augmented training trajectories.
For the former, instead of conditioning the lower-level diffusion models purely on the final output of the high-level diffusion model we also use \emph{intermediate outputs} from the higher-level diffusion model.
These intermediate outputs are noisy version (with random Gaussian noise) of the clean outputs. Hence, they make the  lower-level diffusion model robust to noise in the reference trajectory points.


\begin{figure}[bt]
    \centering
    \includegraphics[width=0.5\textwidth]{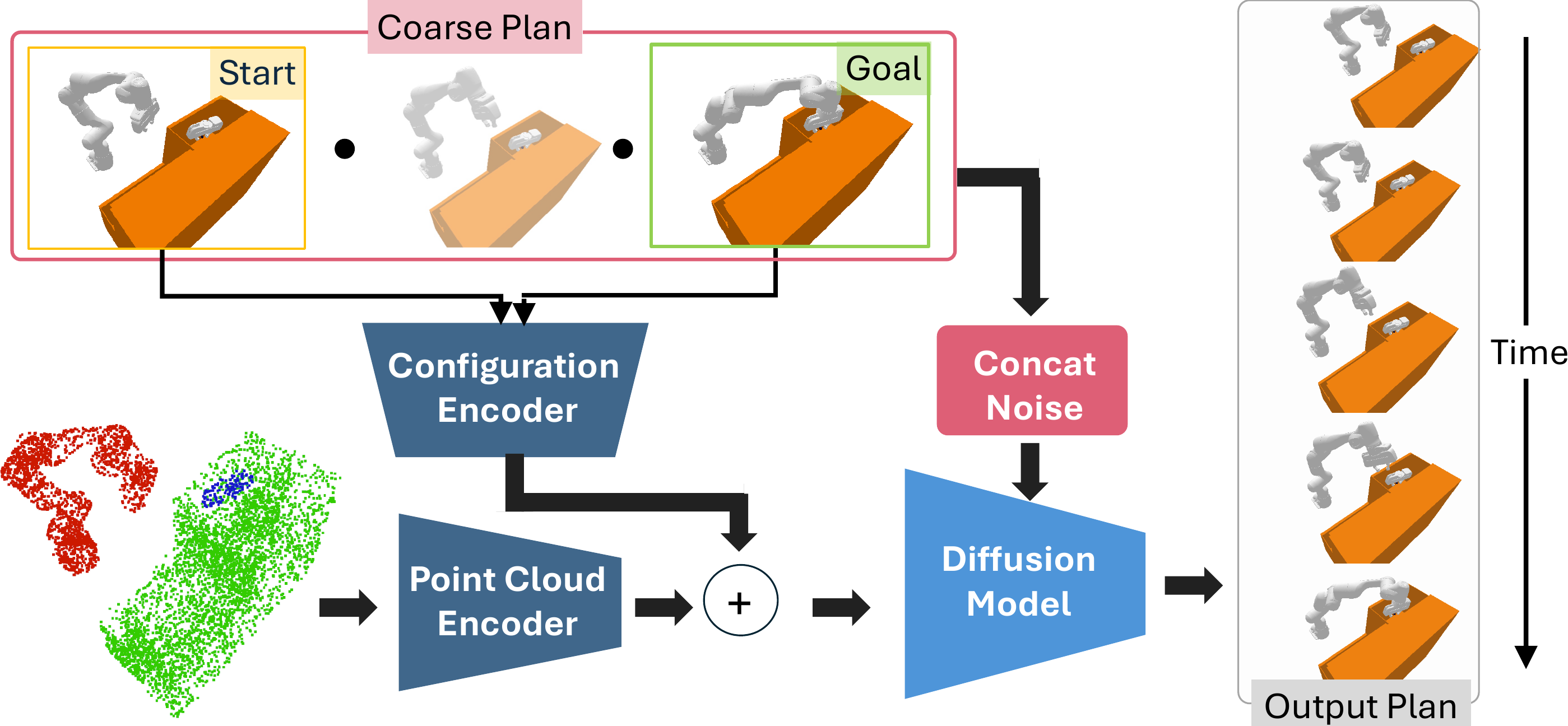}
    \caption{\footnotesize{Our cascaded diffusion model architecture uses high-dimensional observations (point clouds) and coarse plans from high-level model to output complete plans using a diffusion process.}}
    \label{fig:architecture-overview}
\end{figure}

\textbf{Architecture Details:}
Fig~\ref{fig:architecture-overview} shows our proposed architecture.
We use raw point cloud observations $o$, to model the scene. We add the robot geometry at configuration $q_0$ into this point cloud
to provide additional context for the encoder.
This point cloud observation is first processed by a PointNet++ \cite{qi2017pointnet++, fishman2023motion} based observation encoder.
We also pass the current configuration of the robot (and sub-goal configurations when they are available) 
through the configuration encoder. 
We concatenate point cloud and configuration representations.
This representation together with the provided reference trajectory points is used to condition our diffusion model.
The diffusion model uses a temporal U-Net \cite{unet} architecture, 
wherein the provided reference trajectory points are used as additional channels together with random Gaussian noise.
The diffusion model outputs the clean version for this Gaussian noise.
Along with the diffusion loss, similar to \cite{fishman2023motion}, we also add a collision loss to the output of the diffusion model.
We only add the collision loss for diffusion steps close to the final stages of the backward process.

\begin{figure}[t]
    \centering
    \includegraphics[width=0.9\linewidth]{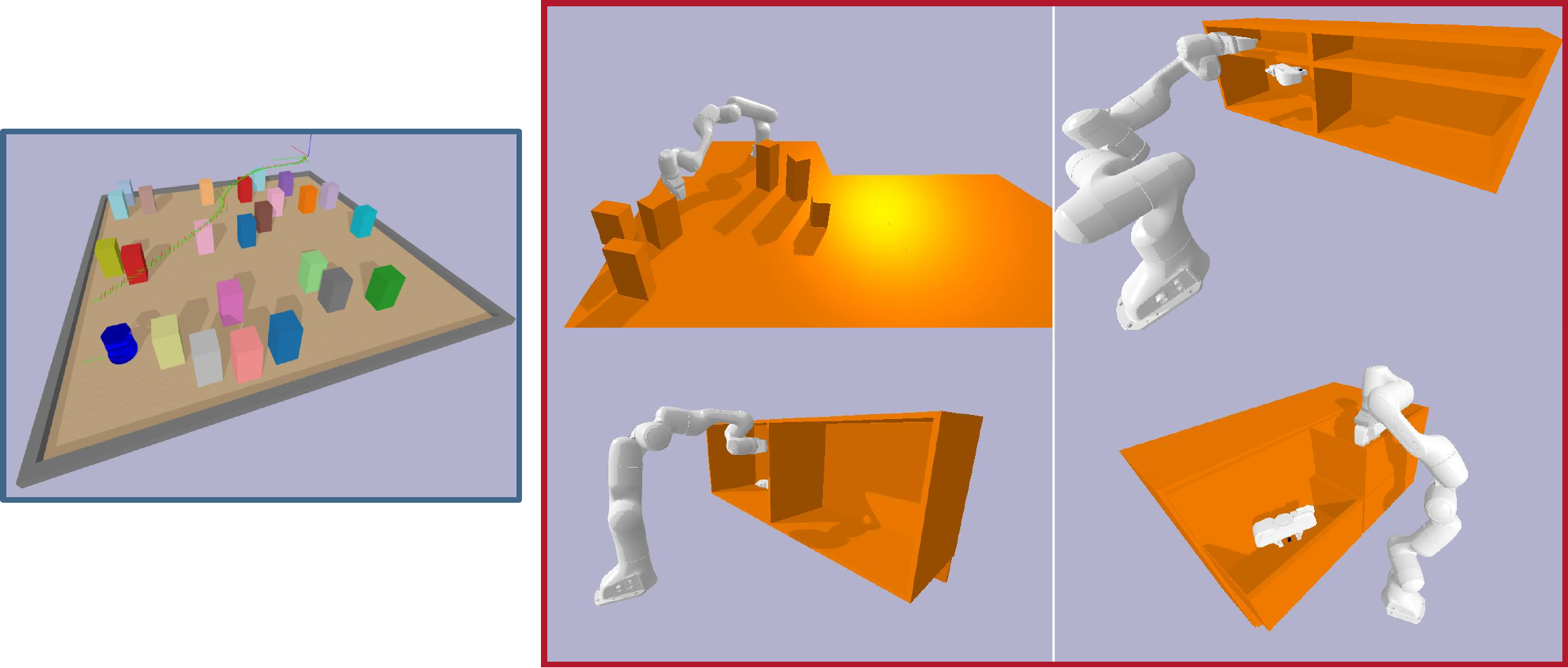}
    \caption{\footnotesize{Overview of the environments we use to evaluate our approach. We use a 2D navigation problem (left) as well as multiple simulated reaching tasks using a simulated Franka Panda arm~\cite{fishman2023motion}.}}
    \label{fig:envs}
\end{figure}

\section{Experiments}

The overall goal of our experiments is to determine:
1) How does our proposed hierarchical diffusion model compare to other learning and non-learning based planning approaches,
and 2) How does our approach perform in scenarios with novel (different) obstacle configurations.

\vspace{-4pt}
\subsection{Experimental Setup}
\label{subsec:exp-setup}
We evaluate our approach
in both 2D and 3D environments, looking at a variety of local and global motion goals.

\textbf{Environments:}
First, we evaluate on a 2D navigation task (Fig~\ref{fig:envs} Left), where we use top-down images as observations.
We generated ground-truth motion plans for this benchmark using RRT~\cite{kuffner2000rrt}, which were used to train all learning-based approaches. We generated $40,000$ train environments and evaluated on $1,000$ held-out test environments.
We also tested on a more complex manipulator setting,
from \mpinetname~\cite{fishman2023motion} (see Fig~\ref{fig:envs} Right).
Specifically, \mpinetname \cite{fishman2023motion} evaluates robot motion planning for a 7-DOF Franka robot in many diverse everyday settings, and uses procedurally generated environment configurations to generate large amounts of diverse training data.
For these environments \mpinetname provides 3D point clouds of the scene as the main observation space.
To generate data \mpinetname uses two different schemes.
First, they provide a \textbf{Global} dataset which is generated by running a global planner (\aitstar) and directly using the data provided by the planner.
They also provide a \textbf{Hybrid} dataset which is generated by first running a global \aitstar \cite{strub2020adaptively} planner in the Cartesian space and then tracking this generated trajectory using a local Geometric Fabrics planner \cite{van2022geometric} to produce smooth and consistent trajectories.
Overall, Global and Hybrid datasets consist of $6.4M$ and $3.2$ million trajectories respectively.
\mpinetname also provides two test sets to evaluate trained planners --
1) the \textbf{Global} test set which consists of 1800 test problems solvable by a global planner and, 2) the \textbf{Hybrid} test set which also consists of 1800 test problems solvable by the hybrid planner.

\textbf{Baselines and Metrics:}
We compare our approach against relevant baselines which focus on using learning for plan generation.
We compare against \mpinetname \cite{fishman2023motion} which propose to learn a fast and reactive policy on motion planning data. 
We also compare against Diffuser \cite{janner2022planning} which uses a diffusion model for planning.
Additionally, we compare against more recent diffusion based approaches which focus on classifier based guidance \cite{carvalho2023motion, saha2023edmp}.
Among them, we use the more recent EDMP \cite{saha2023edmp} which uses an ensemble of differentiable classifier costs to guide the diffusion model.
We also compare against a hierarchical diffusion approach \cite{chen2024simple} which \emph{independently} trains two different diffusion models at different hierarchies.
Further, \cite{chen2024simple} directly uses the sub-goals generated by the higher-level diffusion model as targets for the low-level diffusion model.

We compare each approach by evaluating how well  it allows the robot to 
move from the start to the goal configuration.
We consider a plan successful \emph{only if} the robot arm reaches close to the goal configuration and if no state in the plan the robot has any environment or self-collisions.
Beside \emph{success} criterion we also measure the wall-clock time for each approach.
For qualitative results please see: \url{https://sites.google.com/view/cascaded-diffusion-planning}


\begin{table*}[]
\centering
\resizebox{1.0\textwidth}{!}{%
\begin{tabular}{@{}llllll|llllll@{}}
\toprule
 & \multicolumn{5}{c|}{Solution Time (seconds) $(\downarrow)$} &  & \multicolumn{5}{c}{Success Rate $(\uparrow)$} \\ \midrule
Env & \mpinetname \cite{fishman2023motion} &  Diffuser \cite{janner2022planning} & EDMP \cite{saha2023edmp} (*) & H Diffuser & Ours & & \mpinetname \cite{fishman2023motion} & Diffuser \cite{janner2022planning} & EDMP \cite{saha2023edmp} (*) & H. Diffuser \cite{chen2024simple} & Ours \\ \midrule
Turtlebot & 0.41 & 0.44 & 0.49 & 0.91 & 1.25 & & 71.0 & 82.4 & 88.16  & 87.40 & 91.20 \\
Franka (Hybrid) \cite{fishman2023motion} & 0.33 & 1.61  & 1.94  & 2.48 & 2.74 &  & 95.33 & 90.45 & 94.43 & 97.4 & 98.0 \\
Franka (Global) \cite{fishman2023motion} & 0.33 & 1.61  & 1.94 & 2.48  & 2.74 &  & 75.06 & 68.14  & 80.3 & 80.71 & 85.13 \\ \bottomrule \\
\end{tabular}%
}
\caption{\footnotesize{Comparison to learning based approaches. (*) indicates that the approach uses privileged information (full scene geometry).}}
\label{tab:learn-to-plan-comparison}
\end{table*}

\begin{table}
\centering
\resizebox{0.5\textwidth}{!}{%
\begin{tabular}{@{}llll|llll@{}}
\toprule
 & \multicolumn{3}{c|}{Solution Time (seconds) $(\downarrow)$} &  & \multicolumn{3}{c}{Success Rate $(\uparrow)$} \\ \midrule
Env & OMPL \cite{kuffner2000rrt} & CHOMP \cite{chomp} &  Ours &  & OMPL \cite{kuffner2000rrt} & CHOMP \cite{chomp} &  Ours \\ \midrule
Turtlebot & 1.40 & 0.08 & 1.25 &  & 100.0 &  46.25  & 91.20 \\
Franka (Hybrid plans) \cite{fishman2023motion} & 7.37  & 0.14 & 2.74 &  & 100.0 & 31.61  & 98.0 \\
Franka (Global plans) \cite{fishman2023motion} & 16.46  & 0.18  & 2.74 &  & 100.0  & 26.67  & 85.13 \\ \bottomrule \\
\end{tabular}%
}
\caption{\footnotesize{Comparison to planning based approaches.}}
\label{tab:planner-comparison}
\end{table}


\begin{table}[t]
\centering
\resizebox{0.5\textwidth}{!}{%
\begin{tabular}{lccc}
\toprule
 & \textbf{No Guidance} & \textbf{Guidance w/ ground truth} & \textbf{Noisy Guidance} \\
\midrule
Ours & 85.13 & \textbf{92.31} & 90.95 \\
Ours (No refinement) & 82.74  & 85.62 &  80.61 \\
\bottomrule \\
\end{tabular}
}
\caption{\footnotesize{Success rate for our proposed cascaded diffusion model on the \mpinetname Global dataset with and without motion refinement under various guidance settings.}}
\label{tab:cascaded-with-guidance}
\end{table}

\vspace{-4pt}
\subsection{Results}
\label{subsec:exp-results}

\textbf{Comparison To Learning-Based Approaches:}
Table~\ref{tab:learn-to-plan-comparison} compares the performance of our approach against learning based planning approaches.
From the above table we see that the single step prediction approach of \mpinetname performs sub-optimally on most environments.
We believe this reduced performance of \mpinetname is due to its use of less expressive MLP architectures. Further, \mpinetname uses a mean-squared loss which makes it challenging to learn from planning data which is often multi-modal.
Among the different diffusion based architectures, 
we find that simply using a diffusion process to model the entire trajectory (as in Diffuser \cite{janner2022planning}) can be challenging.
Interestingly, Diffuser \cite{janner2022planning} performs worse ($\approx -5\%$ on average) than 
\mpinetname on the Franka datasets. 
This is because Franka datasets require motion generation for a high-dimensional system in tight constrained spaces and most often the 
plan generated by Diffuser violates the environment collision constraints.

While Diffuser \cite{janner2022planning} performs sub-optimally, other diffusion based architectures improve upon its performance. 
EDMP \cite{saha2023edmp},  which uses an ensemble of more than 10 different cost functions performs better than \mpinetname.
However, EDMP's reliance on multiple cost functions, each with its own set of hyper-parameters,
makes it challenging to scale.
Further, it uses \emph{ground-truth geometry} to 
estimate collision gradients during the guidance process.
This can be problematic since estimating ground truth geometry from noisy point clouds can be challenging.
By contrast, hierarchical diffusion-based approaches (either with independent models~\cite{chen2024simple}, or our approach) outperform all other approaches without relying on guidance. 
Among the hierarchical approaches, our cascaded approach performs better $\approx 3\%$ averaged over all datasets, while being $\approx 4\%$ better on the challenging \mpinetname Global dataset.

We also compare each approach based 
on the amount of time to generate the plan. 
Along this axes, \mpinetname outperforms all other methods. 
This is because unlike diffusion based approaches \mpinetname does not involve an iterative refinement loop.
Among diffusion based approaches, hierarchical approaches which call the diffusion process multiple ($K$) times have the largest planning time cost.

\textbf{Comparison To Non-Learning Approaches:}
Table~\ref{tab:planner-comparison} compares the performance of our approach against common planning based approaches across all different task configurations.
From Table~\ref{tab:planner-comparison} we see that local planning based approaches such as CHOMP \cite{chomp} often perform sub-optimally.
This is because the considered environments are difficult and since local approaches rely on good initialization, 
they are often insufficient to successfully reach the goal. 
On the other hand, sampling based planning approaches \cite{kuffner2000rrt, sucan2012the-open-motion-planning-library} perform quite well, most often reaching the desired goal configurations.
However, due to their reliance on collision checking and naive sampling, the overall wall-clock time for finding a plan is much larger.
By comparison, our proposed learning-based approach performs well across all different environment configurations while simultaneously being fast. 

\begin{figure}[t]
  \centering
  \includegraphics[width=.99\linewidth]{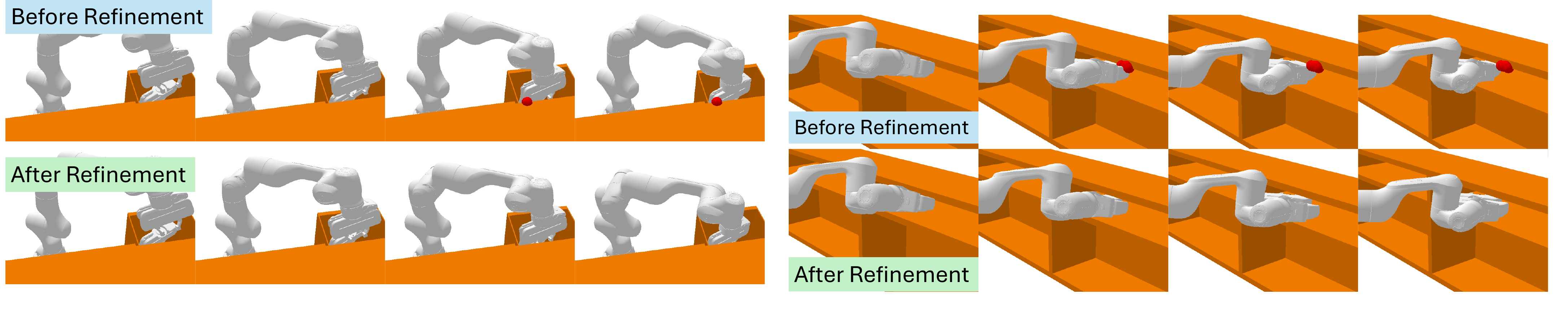} 
  \caption{\footnotesize{Qualitative results. \textcolor{red}{Red} spheres show collisions with the environment. Note that many collisions are very subtle, and the optimal solution is very close to being in collision; this is part of the difficulty of our problem setting.}}
  \label{fig:qualitative-results}
\end{figure}

\begin{figure}{h}
    \centering
    \includegraphics[width=0.9\linewidth]{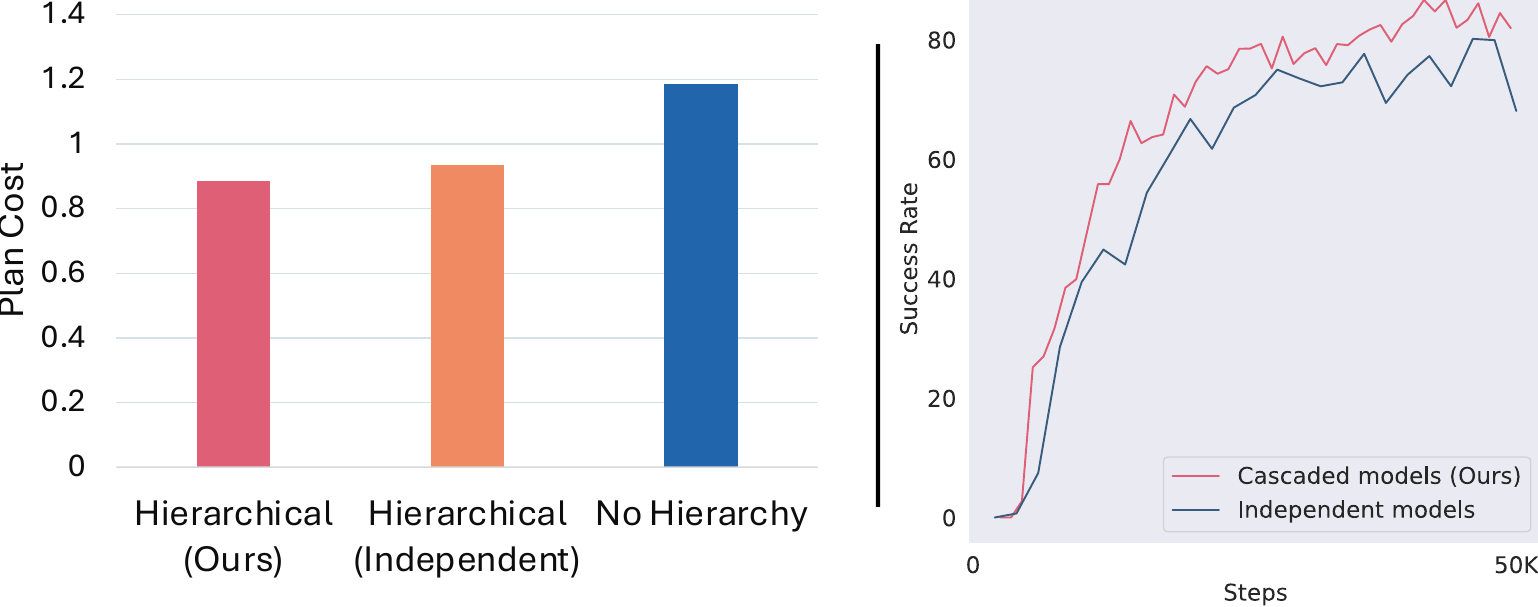}
    \caption{\footnotesize{Planning performance of hierarchical approaches.}}
    \label{fig:plan-cost}
\end{figure}

\textbf{Plan Performance:}
Fig~\ref{fig:plan-cost} compares the most successful learning based approaches (i.e. 
hierarchical approaches) in terms of planning cost and learning efficiency.
Fig~\ref{fig:plan-cost} (Left) plots the planning cost of both approaches and the non-hierarchical approach on the Global dataset.
From this figure we see that hierarchical approaches output plans with lower plan costs
compared to the non-hierarchy-based approach of Diffuser \cite{janner2022planning}.
However, both hierarchical models, i.e. separately trained independent models and our cascaded approach, have a similar cost performance. 
Additionally, Fig~\ref{fig:plan-cost} (Right) compares the success rate of these hierarchical models.
We see that using reference trajectories provides improved performance as well as learning efficiency compared to independently trained models. 
This is because the low-level model in our cascaded approach is able to utilize the output from higher level models to bootstrap learning.

\textbf{Robustness Analysis:}
To assess the robustness of our approach we
analyze the success rate of the model without the path refinement step.
Further, we also evaluate performance with additional guidance (similar to EDMP) for the high-level model \emph{only}.
Table~\ref{tab:cascaded-with-guidance} reports success rate for our approach in the above
settings on the \mpinetname \emph{Global} dataset.
As seen above (column 1), both our cascaded approach and the path refinement contribute almost equally ($\approx 2\%)$ to the overall performance.
Additionally with collision based guidance (middle-column) 
our approach is able to further improve its performance to $92.3\%$ ($+7\%$).
Since, we only add guidance to our high-level model this shows that better sub-goals and reference trajectories significantly improve the overall performance of our approach.

However, as noted in Section~\ref{subsec:exp-results}, EDMP assumes full scene geometry to estimate collision guidance gradients.
Since getting such geometry may be challenging in certain application domains, we also evaluate the performance with noisy gradients (last column) in which we add significant perturbations to the true gradients to mimic real-perception noise.
Importantly, we see that without path refinement the performance with noisy guidance drops ($\approx -5\%$).
By contrast, with additional path refinement, noisy guidance has a much lesser impact $(\approx -1.4\%)$.
This resilience is because the refinement step is trained to utilize noisy reference trajectories and sub-goals to predict optimal paths.
Hence, its performance does not deteriorate even with noisy guidance from the above diffusion process.
Overall, the results shows the robustness of our 
approach in realistic settings.

Finally, Figure~\ref{fig:qualitative-results} shows qualitative results of our path refinement across two different test environments in \mpinetname Global test set.
As can be seen in the above figure,
the goal configurations are often extremely close to 
scene objects. 
This requires the model to carefully plan its motion which 
the high-level model often fails to achieve.
By contrast, our cascaded approach with its local low-level diffusion model and path refinement is able to output plans that satisfy such hard local constraints directly from raw perceptual input.

\section{Conclusion \& Limitations} 

In this work we propose a hierarchy of cascaded diffusion models to generate 
motion plans in complex manipulation environments.
We show that using a cascaded hierarchy of models allows us to better capture 
both global and local information.
Using global information allows our model to generate optimal-cost plans, 
while local information allows it to satisfy the challenging
constraints that occur in high dimensional problems.
We show how our approach is able to outperform all prior methods (while simultaneously being robust) across 3 different environment configurations.
Our work has several limitations. 
First, our approach may struggle in completely novel out-of-distribution scenarios.
While large scale imitation learning alleviates it to some extent,
we found transfer to very different scenes (with novel object geometry) to be challenging.
We believe using our diffusion model's output to bootstrap motion planner sampling can provide better generalization.
Another limitation of our approach is that the planning time is still substantially large ($\approx 1-2$) seconds.
This is due to the iterative refinement within diffusion models.
Using 
recent consistency based diffusion models,
which provide fast sampling with few diffusion steps  may alleviate this \cite{li2024self, song2023consistency}.


\clearpage

\bibliographystyle{IEEEtran} 
\bibliography{IEEEabrv,references}

\end{document}